\documentclass[lettersize,journal]{IEEEtran}

\IEEEoverridecommandlockouts                              

\usepackage{booktabs}
\usepackage{array}
\usepackage{multirow}
\usepackage{makecell}
\usepackage{graphicx} 
\usepackage{adjustbox}
\usepackage{graphicx} 
\usepackage{subcaption}
\usepackage{tabularx}
\usepackage{caption}
 \usepackage{url}
\usepackage{xcolor}
\definecolor{codebg}{gray}{0.96}
\usepackage{listings}
\usepackage{newfloat}
\DeclareFloatingEnvironment[name=Listing]{listing}
\newif\ifdraft
\drafttrue      

\ifdraft
  \newcommand{\todo}[1]{\textcolor{red}{\textbf{TODO:} #1}}
\else
  \newcommand{\todo}[1]{} 
\fi

\title{\LARGE \bf 
BiPOP: Partial-Order Planning for Long-Horizon Bimanual Robotic Manipulation
}

\author{
{Kun Chu}$^{*}$, {Xufeng Zhao}, {Jan-Gerrit Habekost}, {Cornelius Weber}, and {Stefan Wermter}
\thanks{The authors are with the Knowledge Technology Group, Department of Informatics, University of Hamburg, 20146 Hamburg, Germany. E-mail:
        {\tt\small \protect\\ {\{kun.chu, xufeng.zhao, jan-gerrit.habekost, cornelius.weber, stefan.wermter\}@uni-hamburg.de}}}
\thanks{$^{*}$Corresponding author.}
}

\begin{document}
\maketitle
\thispagestyle{empty}
\pagestyle{empty}


\begin{abstract}
Bimanual robotic manipulation offers substantial versatility, yet its central challenge lies not only in executing dexterous motion with each hand, but also in coordinating the two arms for long-horizon manipulation under coupled spatial and temporal constraints. Prior work has largely focused on endowing robots with human-level manipulation skills, paying comparatively less attention to the structure of bimanual behaviour itself: many practical tasks are best understood as a coordinated composition of reusable unimanual skills together with a small set of synchronous bimanual skills. This makes high-level spatio-temporal coordination, rather than skill acquisition alone, a key bottleneck for efficient bimanual execution. Therefore, we propose BiPOP, a framework that couples Large Language Models (LLMs) with Temporal Partial-Order Planning for long-horizon bimanual manipulation. At its core is an explicit spatio-temporal model that partitions the workspace by arm reachability and captures inter-arm dependencies, arm assignment, synchronization requirements, and opportunities for safe concurrent execution. Building on this model, an LLM translates natural-language instructions into a formal PDDL specification, while a temporal planner enforces the constraints and schedules actions, yielding plans that are logically consistent at the symbolic level and efficient through parallel execution. On simulated long-horizon tasks of varying complexity, BiPOP outperforms open-loop planning with strong reasoning models and closed-loop ReAct-style planning in success rate, planning time, and plan efficiency, and the full pipeline runs end-to-end on a real bimanual robot NICOL. A failure analysis further shows that LLM-only planners mainly violate spatial and temporal coordination constraints, which motivates delegating these constraints to a symbolic planner. Our implementation is publicly available at \url{https://github.com/Kchu/BiPOP}.
\end{abstract}

\section{Introduction}
Humans effortlessly perform bimanual tasks, from tying shoelaces to flipping book pages while holding a coffee cup, orchestrating complex spatial and temporal synchronization without conscious thought. Robots, in contrast, struggle even with apparently simple bimanual tasks, which demand planning, synchronization, and coordination. Bridging this gap remains an inherent challenge in robotics \cite{Krebs2022, Drolet2024}. 

Deep learning techniques ease the acquisition of dexterous bimanual manipulation, such as folding shirts \cite{shawbimanual}, cutting vegetables \cite{Grannen2023}, and scooping food \cite{Grannen2023a}. However, most approaches focus on learning human-level skills as a whole and overlook high-level task planning, which demands task decomposition, subtask assignment between the arms, execution sequencing and parallelization over long horizons.

Large Language Models (LLMs) have shown strong reasoning and planning abilities across domains \cite{zhao2023survey}, and recent work applies them to robotic task planning \cite{Ahn2022, liang2023code, Chu2024, gao2024dag}. Yet generating embodiment-aware, executable plans is difficult for complex manipulation, where LLMs remain prone to long-horizon reasoning errors and hallucinations \cite{wang2024survey, stechly2024self}, especially in spatially constrained scenarios. To mitigate these limitations, LLMs have been combined with classical symbolic planners \cite{liu2023}: by translating natural-language instructions into declarative representations such as PDDL \cite{mcdermott1998pddl}, LLMs enable plans with logical-correctness guarantees. However, these approaches generally do not address the spatio-temporal coordination intrinsic to bimanual manipulation, in particular inter-arm dependencies, task allocation, synchronization, and parallelization. Our research question is therefore: \textit{how can LLMs be integrated with symbolic planning to achieve efficient spatial and temporal coordination for long-horizon bimanual robotic tasks?}

\begin{figure*}[ht!]
 \begin{center}
  \centerline{\includegraphics[width=1\textwidth]{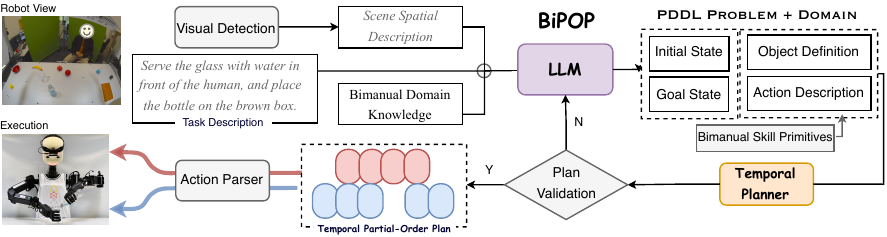}}
 \end{center}
 \vspace{-0.7cm}
 \caption{Overview of BiPOP, illustrated on the ServeWater task. From the visual scene description, the bimanual domain knowledge, and the natural-language instruction, BiPOP generates a PDDL problem solved by a temporal planner. The validated partial-order plan with the best symbolic makespan found (cf.~Fig.~\ref{fig:temporal_partial_order_plan}) is executed by the bimanual robot NICOL~\cite{Matthias2023}.}
\label{fig:method}

\end{figure*}

To address this, we propose \textbf{BiPOP} (Bimanual Partial-Order Planning), a framework (Fig. \ref{fig:method}) for long-horizon bimanual task planning that integrates semantic reasoning with explicit spatio-temporal coordination. Our contributions are threefold. First, we introduce a spatio-temporal model that partitions the workspace by arm reachability and encodes inter-arm dependencies, synchronization, and safe concurrency as durative PDDL actions with mutual-exclusion predicates. Second, we couple LLM-based problem generation with temporal planning: an LLM translates the task description and scene information into a PDDL problem, and a temporal planner computes a validated partial-order plan with a short symbolic makespan. Third, we evaluate BiPOP against open-loop and closed-loop LLM planners on three simulated bimanual task domains and deploy it on a real bimanual robot NICOL~\cite{Matthias2023}. The results show that BiPOP outperforms open-loop planning with recent OpenAI and DeepSeek reasoning models and closed-loop ReAct-style planning in success rate, planning time, and plan efficiency.

\section{Related Work}\label{sec:related}
\subsection{Task Planning in Robotics}

\noindent\textbf{Planning with Symbolic Planners.} Task planning generates action sequences that achieve specified goals \cite{gerevini2020introduction, jiang2019task}. Classical planners formalize domains and problems in declarative languages, most prominently PDDL \cite{mcdermott1998pddl}, and generate plans with guaranteed logical correctness through domain-independent search \cite{helmert2006fast, Eyerich2009}. Task and motion planning couples such a symbolic planner in discrete space with a motion planner in continuous space \cite{Ding2020, kaelbling2013integrated, jiao2021efficient}. These planners provide formal validity guarantees but require expert effort to author the symbolic representations, motivating the use of LLMs to automate this step.

\noindent\textbf{Planning with LLMs.} Prompting LLMs to generate robot behaviour directly is now a common technique \cite{Ahn2022, Chu2024, huang2022language, ding2023task}. SayCan \cite{Ahn2022} lets an LLM propose feasible skills and grounds them to the current scenario through learned value functions, while language-to-code methods ground language more directly in executable robot behaviour: Code as Policies \cite{liang2023code} writes policy code over perception and control primitives, and VoxPoser \cite{huang2023voxposer} composes LLM/VLM-generated 3D value maps for zero-shot trajectory synthesis. These methods turn short-horizon instructions into executable, mostly single-arm motion, but presuppose a short or given task-level sequence and execute the LLM output unverified. Although iteratively querying LLMs with feedback or error messages mitigates these issues \cite{Chu2024, Wang2023, Rana2024}, such strategies still offer no formal guarantee of plan validity and rely on hand-crafted, domain-specific feedback with considerable expert effort.

To pair language understanding with formal guarantees, a growing line of work uses LLMs not as planners but as translators of natural-language tasks into PDDL solved by classical planners. LLM+P \cite{liu2023} established this paradigm; ViLaIn \cite{shirai2024vision} adds visual grounding, DELTA \cite{liu2025delta} decomposes long-horizon goals into sub-goals, and AutoGPT+P \cite{birr2024autogptp} handles incomplete information via affordance-based scene representations. However, these methods plan for a single agent and produce strictly sequential plans. Although AutoGPT+P is validated on bimanual humanoid platforms, it models neither the two arms as coordinated agents nor the action durations and concurrency constraints required for bimanual manipulation. LLM+MAP~\cite{chu2025llmmap} combines LLM reasoning with PDDL-based multi-agent planning for bimanual manipulation. Although it also produces partial-order plans, it frames coordination primarily as task and resource allocation between the two arm agents, rather than explicitly optimizing makespan through temporal planning. 


\begin{figure*}[th]
 \begin{center}
  \centerline{\includegraphics[width=1\linewidth]{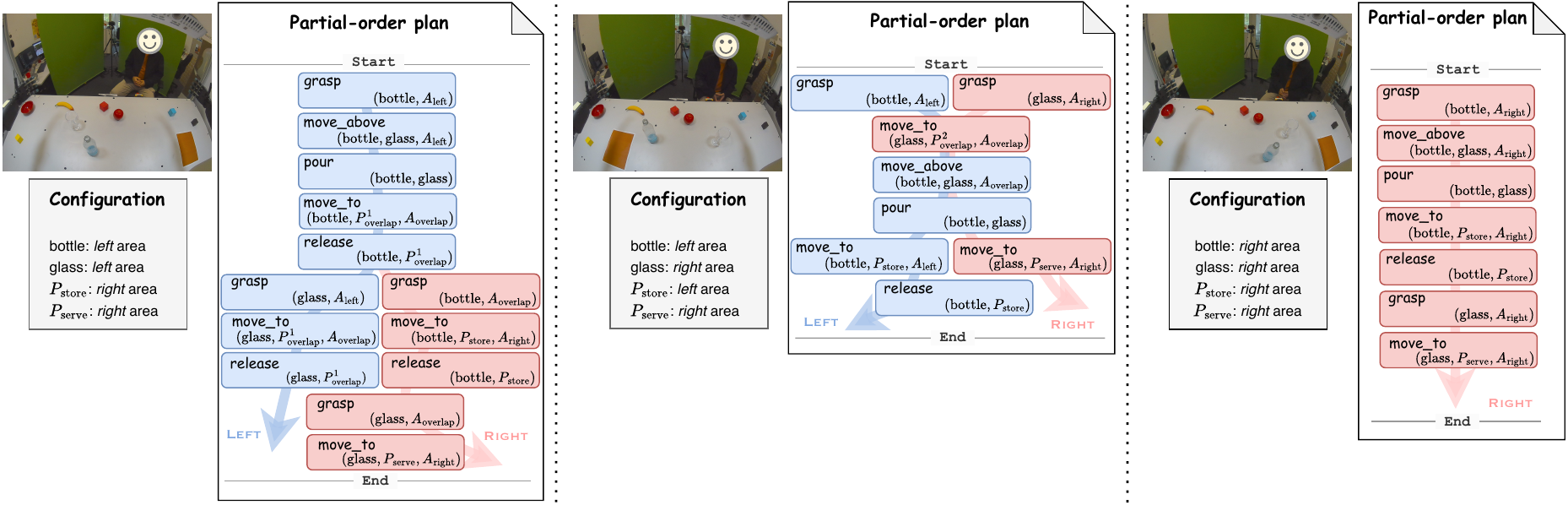}}
 \end{center}
    \vskip -1.5em
     \caption{Temporal partial-order plans generated by BiPOP for the ServeWater task on the real robot NICOL under three different initial scene configurations. $A$ and $P$ denote \textit{area} and \textit{point}, respectively. Light-blue and red boxes represent left- and right-arm durative actions, while horizontally aligned boxes indicate actions executed in parallel. }
    \label{fig:temporal_partial_order_plan}
    \vspace{-0.4cm}
\end{figure*}

\subsection{Bimanual Manipulation and Planning}

\noindent\textbf{Bimanual Manipulation.}
Deep learning has enabled dexterous bimanual skills \cite{Drolet2024} with parallel grippers \cite{Grannen2023} or anthropomorphic hands \cite{Chen2022}, from scooping food \cite{Grannen2023a} to folding clothes \cite{avigal2022speedfolding} and zipping zippers \cite{Grannen2023}. As these works focus on skill acquisition, they do not address planning for complex, long-horizon tasks; the diversity of coordination patterns they exhibit suggests that high-level planning should be considered jointly in an integrated control-system design \cite{Krebs2022, Smith2012}.

\noindent\textbf{Bimanual Planning.}
Recent bimanual planning methods fall into two main categories. (i) \textit{Multi-agent formulations} \cite{Zhao2024, zhao2025dual} model each arm as an autonomous agent and use LLMs to decompose and allocate subtasks. This design is flexible, but typically lacks explicit concurrency scheduling, so coordination rests on heuristic LLM reasoning rather than verified constraints. (ii) \textit{Graph-centric pipelines} \cite{gao2024dag, duan2026robopara} induce a task-dependency graph with an LLM and dispatch its nodes to the two arms, exploiting parallelism where precedence permits. DAG-Plan \cite{gao2024dag} assigns nodes to arms reactively at run-time via heuristic feasibility and cost estimates, leaving the LLM-generated dependencies unverified. RoboPARA \cite{duan2026robopara} validates its graph only against a few hand-crafted structural error patterns and schedules it greedily with deadlock rollback, without a correctness guarantee or an explicit makespan objective. In both, the dependency semantics depend substantially on LLM assertions. In contrast, we ground concurrency in a verified symbolic model: the workspace partition and skills are formalized as temporal PDDL, and a temporal planner determines which actions may remain partially ordered and which require coordination, minimizing makespan with correctness guaranteed at the symbolic level.
 
\section{Method}\label{sec:method}

Given a task description in natural language, we aim to generate a valid and efficient plan for a bimanual robot based on the initial scene configurations, e.g., spatial information about the objects with respect to the robot. To this end, this section introduces the three elements of \textbf{BiPOP}: {\em (A)} Spatio-temporal modelling, {\em (B)} \textbf{Bi}manual PDDL definition with the LLM, and {\em (C)}~\textbf{P}artial-\textbf{O}rder \textbf{P}lanning with the temporal planner.

\subsection{Spatio-temporal Modelling} 
\begin{figure}[htb]
 \begin{center}
  \centerline{\includegraphics[width=0.8\columnwidth]{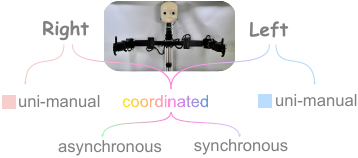}}
 \end{center}
     \vspace{-0.6cm}
     \caption{Spatio-temporal modelling between the two hands.}
    \label{fig:category}
    \vspace{-0.1cm}
\end{figure}

To enable efficient bimanual manipulation, we first develop a spatio-temporal model that characterizes the reachable workspace of each arm and the coordination relations between them, as illustrated in Fig.~\ref{fig:category}. The workspace is partitioned into two uncoordinated regions, each reachable by a single arm, and one coordinated region (the \textit{overlap area}) reachable by both. Arms operating in separate uncoordinated regions on distinct objects are spatially independent and can execute concurrently as parallel unimanual skills. Within the coordinated region, the arms become spatially and temporally coupled, requiring either asynchronous coordination, in which one arm establishes the preconditions for the other, or synchronous coordination, in which both arms execute a tightly coupled bimanual action.

This workspace partitioning operates at the task-planning level: it determines whether two actions may proceed concurrently based on the arms, objects, and regions they involve, rather than on the geometric feasibility of the underlying trajectories. Two actions are scheduled concurrently only when they share no symbolic resource (neither arm, object, nor region), while all remaining cases are serialized through explicit coordination (ordering) constraints. Consequently, the resulting schedules are sound at the task level. Collision-free execution of concurrent motions, including cross-arm trajectory interference near the coordinated region, is delegated to the downstream motion planner. When a concurrent assignment proves geometrically infeasible, the motion planner may serialize it without affecting the validity of the produced plan.

Planning for a bimanual task thus begins with identifying the regions in which the task-relevant objects and target poses lie: an open-vocabulary object detector (e.g., LocateAnything \cite{wang2025locateanything}) localizes the objects, and the resulting bounding boxes are mapped onto the workspace partition.

\subsection{Bimanual PDDL Definition with the LLM}
\subsubsection{PDDL Formalization}

The Planning Domain Definition Language (PDDL) \cite{mcdermott1998pddl} is a declarative language for formalizing planning problems in a standardized form. A planning problem is defined as a tuple $\mathcal{T} = \langle \mathcal{O}, \mathcal{P}, \mathcal{A}, \mathcal{S}, \mathcal{G}\rangle, $ where $\mathcal{O}$ is the set of objects, $\mathcal{P}$ the set of predicates, $\mathcal{A}$ the set of actions, $\mathcal{S}$ the initial state, and $\mathcal{G}$ the goal. A state is a set of grounded predicates, instantiated over objects with Boolean values, and the goal is a logical expression over such predicates; each action is specified by typed parameters together with preconditions and effects expressed as predicates over these parameters. A PDDL specification comprises a \textit{domain}, declaring the object types, predicates, and parameterized actions of the task world, and a \textit{problem}, instantiating a concrete set of objects, an initial state, and a goal.
\subsubsection{Domain Definition}
To encode the proposed spatio-temporal model and the underlying manipulation logic, we define four object types, \textit{arm}, \textit{area}, \textit{object} (\texttt{obj}), and \textit{point}, together with predicates describing their properties and relations. To support the durative actions introduced below, we additionally define predicates that represent the execution state. The predicate \texttt{available} indicates whether an arm is unoccupied, e.g., not holding an object, while \texttt{arm\_free} indicates whether it is currently idle, which prevents a single arm from executing multiple actions at once. The predicate \texttt{is\_static} indicates whether an object is at rest rather than being moved. These predicates encode mutual-exclusion conditions that prevent infeasible concurrent actions, such as two actions manipulating the same object simultaneously, while still permitting independent actions to run in parallel.
For temporal planning, we model each action as a durative action, distinguishing predicates that must hold at the start of an action, throughout its execution, and at its end, as shown in Listing~\ref{lst:pddl_action}. This temporally annotated form is essential for expressing the overlap and ordering constraints that arise when two arms act concurrently.

We design a set of atomic skills for the two identical robotic arms, covering both independent single-arm operations and joint dual-arm operations. Each symbolic action is grounded in a robotic controller through its arm parameters: a single-arm action takes a specific arm as a parameter and maps directly to that arm's controller, whereas a joint operation is not parameterized by an individual arm. Instead, its precondition requires both arms to be in the \texttt{available} and \texttt{arm\_free} states, and the skill is bound to both arms simultaneously, coupling the two controllers into a single synchronized execution. Region membership enters the domain directly through preconditions such as \texttt{arm\_access(?arm,\,?area)} (cf.\ Listing~\ref{lst:pddl_action}), so reachability constrains action applicability rather than being checked post hoc. Location-dependent skills such as \texttt{pour} carry no explicit target parameter: they act through the \texttt{above} relation established by a preceding \texttt{move\_above} action, as visible in the plans of Fig.~\ref{fig:temporal_partial_order_plan}.

\begin{listing}[t]
\vskip 0.5em
\caption{Durative PDDL actions \texttt{grasp} and \texttt{co\_move\_to}.}
\label{lst:pddl_action}
\begin{minipage}{\dimexpr\linewidth-1.2em\relax}
\begin{lstlisting} 
(:durative-action grasp
  :parameters ( ?x - obj ?y - arm ?z - area)
  :duration (= ?duration 1)
  :condition (and 
    (at start (arm_access ?y ?z))
    (at start (arm_free ?y))
    (at start (available ?y))
    (...))
  :effect (and 
    (at start (not (arm_free ?y)))
    (at start (not (available ?y)))
    (at end (arm_free ?y))
    (...))
        
(:durative-action co_move_to
  :parameters ( ?x - obj ?y_0 - point)
  :duration (= ?duration 1)
  :condition (and 
    (at start (arm_free left_arm))
    (at start (arm_free right_arm))
    (...))
  :effect (...))
\end{lstlisting}
\end{minipage}
\vspace{-0.8em}
\end{listing}

\begin{figure*}[!ht]
    \centering
    \includegraphics[width=0.91\linewidth]{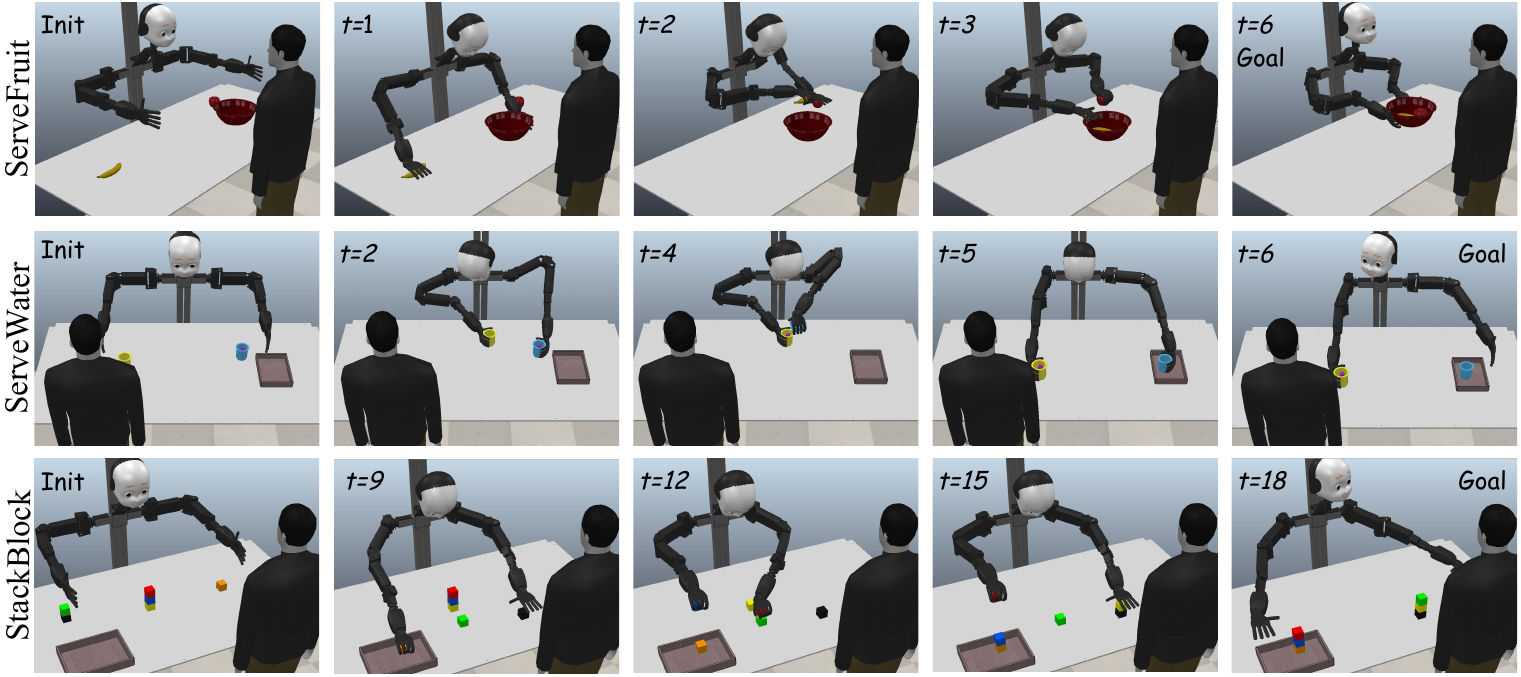}
    \caption{The three task domains in the simulated human-robot scenario. (a) ServeFruit: the human faces the overlap area to receive the bowl; the fruits and the bowl are initialized at random positions. (b) ServeWater: the brown box is in the left or right area to receive the blue cup; the cups and the human are placed randomly. (c) StackBlock: the blocks start randomly placed or stacked, the brown box occupies a random position across the three areas, and the human faces a random area.}
    \label{fig:task_domain}
\end{figure*}

\subsubsection{LLM as a PDDL Writer}

Following the formalization above, we prompt the LLM to generate the \textit{problem} specification: the initial state is derived from the scene description obtained from the camera image, and the goal is derived from the natural-language task instruction. We employ the LLM for this step because, paired with open-vocabulary detection, it provides autonomous visual grounding of the task configuration: free-form object references in the instruction are resolved directly against detected labels. Rather than generating raw PDDL text, we instruct the LLM to produce Python code that constructs the problem through the Unified-Planning library \cite{unified_planning_2025}, which standardizes problem definition and provides a uniform interface to built-in symbolic solvers; this exploits the code-generation ability of LLMs while constraining their output to a well-defined API. To handle generation errors, we re-prompt the LLM with feedback from two sources. When code execution raises an exception, the Python interpreter error message is fed back for regeneration, which corrects syntactic and API-level mistakes. When the code executes successfully but the planner returns no plan within the fixed timeout, this no-plan outcome is likewise returned to the LLM, prompting it to revise the initial state or goal specification. At our task scale, a no-plan outcome almost always stems from an erroneous specification rather than from search timeout, making it a reliable re-prompting signal.

\subsection{Temporal Partial-Order Planning}

We adopt Temporal Fast Downward (TFD) \cite{Eyerich2009} as the temporal planner. TFD supports durative actions with start and end conditions and effects, and it schedules independent actions concurrently, which shortens the makespan while preserving temporal consistency. Applied to our bimanual domain, TFD yields a temporal partial-order plan in which independent unimanual actions run in parallel and coordination-dependent actions are ordered accordingly (cf. Fig.~\ref{fig:temporal_partial_order_plan}).
To exploit the planning budget, we run TFD under several parameter configurations and search strategies, yielding multiple candidate plans of varying quality. While the problem-generation stage uses interpreter and planner feedback to obtain a solvable specification, this stage uses the VAL plan validator\footnote{\url{https://github.com/KCL-Planning/VAL}} as the safeguard of the planner itself: TFD occasionally emits schedules that violate durative-action conditions or the temporal semantics of the domain, most often among the shorter candidates, and VAL rejects these before execution. Among the validated candidates, we dispatch the one with the shortest makespan to the robot controllers.

Our framework operates at the task-planning level, reasoning about the ordering and concurrency of symbolic actions rather than the kinematics of their execution. We therefore assign all actions a uniform duration, since exact skill execution time depends on trajectory length, robot dynamics, and low-level control resolved by the motion-planning layer. This abstraction isolates the effect of concurrency structure from low-level execution variability. Under unit durations, the makespan of a plan equals its number of parallel steps, with concurrent actions counted once. We refer to this quantity as the symbolic makespan and use it as the single efficiency objective throughout the paper. As TFD runs under a fixed time budget, the returned plan has the best symbolic makespan found within this budget rather than a certified optimum. The uniform-duration choice does not restrict the formulation itself: TFD can directly incorporate per-action or state-dependent durations, and integrating motion-layer duration estimates is left for future work.

\section{Experiments}
We compare BiPOP, built on GPT-4.1, against an open-loop and a closed-loop LLM-based planning baseline, I/O and ReAct. I/O generates the entire plan in a single pass and then executes it, whereas ReAct interleaves reasoning and acting, emitting one action at a time and receiving environmental feedback before deciding the next. As the two baselines place different demands on the underlying model, we pair each with an appropriate choice. Open-loop synthesis is reasoning-intensive, since the full plan must be produced at once; we therefore equip I/O with two strong reasoning models, OpenAI o3 (denoted o3) and DeepSeek-R1-0528 (denoted R1), and additionally report I/O with GPT-4.1, the backbone of BiPOP. ReAct, by contrast, re-queries the model at every step, making per-step inference latency critical for a usable closed-loop system; we therefore use GPT-4.1, which offers low-latency responses and matches the backbone of BiPOP. 

This configuration isolates the planning mechanism along two axes: against I/O, BiPOP is evaluated against reasoning models and its own backbone, so no advantage can be attributed to model capability; against ReAct, the backbone is held fixed at GPT-4.1, so the remaining difference reflects the mechanism alone. We further align the prompts across methods. All receive the same task description, scene information, and skill set, differing only in how these are encoded:
\begin{itemize}
\item \textbf{I/O}: the task description (with spatial scene understanding), the spatio-temporal bimanual modelling paradigm, and the skill specifications as natural-language preconditions and effects.
\item \textbf{ReAct}: the same content as I/O, but generated step by step with an observation fed back after each action; if a step for two arms cannot be fully executed, the executable portion is carried out while the remainder and its error message are returned to the agent.
\item \textbf{BiPOP}: the same task description with the general PDDL domain definition and predicate specifications used by the temporal partial-order planner.
\end{itemize}

For I/O and ReAct, the LLM is prompted to generate a valid plan with the fewest parallel steps, i.e., the smallest symbolic makespan, matching the objective of BiPOP. ReAct is additionally given a fixed step budget per task domain. We do not compare against language-to-code policies such as Code as Policies or VoxPoser, which, as discussed in Sec.~\ref{sec:related}, synthesize motion-level behaviour and presuppose a short or given task-level sequence, precisely the capability our baselines are designed to probe.

\subsection{Environment Setup}
For quantitative evaluation, the experiments are conducted in the CoppeliaSim simulator on a bimanual robot with two arms of eight degrees of freedom each, terminated by adult-sized, five-fingered hands for precise handling of everyday objects. Visual perception uses one of the two cameras at each eye's position.
We evaluate BiPOP against the baselines across three task domains within a human-robot environment: ServeFruit, ServeWater, and StackBlock (cf.\ Fig.~\ref{fig:task_domain}), which resemble real-world interactions among the robot, a human, and the task objects.
\noindent\textit{ServeFruit} contains a banana, an apple, and a large red bowl; the robot must place the fruit into the bowl and serve the bowl to the human. As the bowl is too large to be grasped by one hand, it requires simultaneous two-hand manipulation, which is only possible in the overlap area.
\noindent\textit{ServeWater} contains an empty yellow cup and a water-filled blue cup on the table; the robot must serve the water in the yellow cup to the human while placing the blue cup at a specified store point on the brown box.
\noindent\textit{StackBlock} contains six coloured blocks, randomly located or initially stacked on the table. The goal is to stack them into one or two piles in a specified order. Each configuration varies in two aspects: the pile locations, one in front of the human, the other on the brown box, and the block order and count per pile in the initial and goal stacks.

Each domain is evaluated on 100 randomized instances, over which Tables~\ref{tab:plan_time} and~\ref{tab:success_rate} report statistics. For every instance, the object configuration is sampled as described above, and the natural-language instruction is instantiated from a per-domain template with the sampled goal (e.g., the object stacking order). Instructions state only the desired outcome, leaving arm assignment, hand-over strategy, intermediate placements, and action ordering unspecified. They are thus not surface paraphrases of the PDDL goal, and the LLM must ground them jointly with the scene description.

\subsection{Metrics}\label{subsec:metrics}
To assess the efficacy and efficiency of the generated plans, we compare our method with the baselines along three metrics:
\begin{itemize}
    \item \textbf{Planning Time (PT).} For the I/O and ReAct agents, the planning time depends solely on LLM inference, whereas for our framework it comprises both the LLM inference time and the TFD planning time. TFD is run as an anytime planner under several search configurations: within a per-domain optimization budget of $1$\,s, $1$\,s, and $25$\,s for ServeFruit, ServeWater, and StackBlock respectively (scaled to search-space size), it keeps searching for shorter plans, and the best validated plan is returned. Reported TFD times \footnote{Although TFD is given a fixed optimization budget, it may terminate earlier when the search space has been fully explored and no further candidate plans remain.} include grounding and validation in addition to search. 

    \item \textbf{Success Rate (SR).} The proportion of runs whose generated plan achieves the task goal upon execution.

    \item \textbf{Optimality Gap (OG).} A measure of planning efficiency relative to the best plan found by any evaluated method. Each run $r$ denotes one task instance evaluated by all methods. Let $s_m(r)$ be the symbolic makespan of method $m$'s plan on $r$, measured in unit-duration steps (concurrent actions count once), and let $\mathcal{M}(r)$ be the set of methods that solved $r$, with $s^{*}(r) = \min_{j \in \mathcal{M}(r)} s_{j}(r)$. For $m \in \mathcal{M}(r)$,
    \begin{equation}
        \mathrm{OG}_m(r) = s_m(r) - s^{*}(r) \geq 0,
    \end{equation}
    where $\mathrm{OG}_m(r)=0$ marks the most efficient method(s) on $r$ and larger values indicate less efficient plans. The resulting $\mathrm{OG}_m$ distributions of all methods are reported in Fig.~\ref{fig:bar_all_tasks}.
\end{itemize}

\subsection{Results}

\noindent\textbf{Planning Time.} As shown in Table~\ref{tab:plan_time}: (1)~GPT-4.1 and BiPOP require substantially less planning time than the reasoning models o3 and R1; (2)~inference time grows with task complexity across all models; and (3)~for BiPOP, the LLM inference time grows only moderately with task complexity, while the TFD planning time is negligible on simpler tasks but dominates on StackBlock, where the larger object and predicate sets expand the search space. Even so, BiPOP remains faster than both reasoning models on every task.

\begin{table}[!ht]
\vspace*{4pt}
\centering
    \caption{Planning Time (s) across the task domains. For BiPOP, LLM inference (code generation) and TFD planning times are listed separately.}
    \label{tab:plan_time}
    \fontsize{7pt}{7pt}\selectfont
    \renewcommand{\arraystretch}{1.5} 
\begin{tabular}{llccc}
\toprule
\textbf{Method} & \textbf{Model} 
& \textbf{ServeFruit} 
& \textbf{ServeWater} 
& \textbf{StackBlock} \\
\midrule

\multicolumn{1}{l}{\multirow{3}{*}{\scriptsize{\textbf{I/O}}}}
  & \scriptsize{R1}      & 227.07 \tiny{±51.19}  & 245.71 \tiny{±54.10}  &  367.57 \tiny{±76.35} \\
\multicolumn{1}{l}{}
  & \scriptsize{o3}      & 32.74 \tiny{±15.39}   & 43.76 \tiny{±34.86}   & 73.20 \tiny{±37.85} \\
\multicolumn{1}{l}{}
  & \scriptsize{GPT-4.1} & 3.42 \tiny{±0.47}     & 2.31 \tiny{±0.95}     & 4.59 \tiny{±2.04} \\
\midrule

\multicolumn{1}{l}{\textbf{ReAct}} & GPT-4.1 & 16.56 \tiny{±5.47} & 24.74 \tiny{±9.70}& 42.46 \tiny{±22.60} \\
\midrule

\multicolumn{1}{l}{\multirow{3}{*}{\textbf{BiPOP}}}
  & \scriptsize{Total} & 6.56 \tiny{±1.49} & 5.95 \tiny{±1.83} & 35.56 \tiny{±9.02} \\
\multicolumn{1}{l}{}
  & \scriptsize{GPT-4.1} & 4.95 \tiny{±1.44} & 4.98 \tiny{±1.73} & 6.65 \tiny{±1.78} \\
\multicolumn{1}{l}{}
  & \scriptsize{TFD} & 1.60 \tiny{±0.29}  & 0.97 \tiny{±0.34}  & 28.91 \tiny{±7.46} \\
\bottomrule
\end{tabular}
\vspace{-0.1cm}
\end{table}

\begin{table}[!ht]
\centering
    \caption{Success Rate (\%) of plan execution across the task domains.}
    \label{tab:success_rate}
    \fontsize{7pt}{7pt}\selectfont
    \renewcommand{\arraystretch}{1.5}
\begin{tabular}{llccc}
\toprule
\textbf{Method} & \textbf{Model}
& \textbf{ServeFruit}
& \textbf{ServeWater}
& \textbf{StackBlock} \\
\midrule
\multicolumn{1}{l}{\multirow{3}{*}{\scriptsize{\textbf{I/O}}}}
  & \scriptsize{R1} & 66 & 82 & 52 \\
\multicolumn{1}{l}{}
  & \scriptsize{o3}         & 98  & 87 & 86 \\
\multicolumn{1}{l}{}
  & \scriptsize{GPT-4.1}    & 25 & 0 & 0 \\
\midrule
\multicolumn{1}{l}{\textbf{ReAct}} & GPT-4.1 & 100 & 45 & 36 \\
\midrule
\multicolumn{1}{l}{\textbf{BiPOP}} & GPT-4.1 & \textbf{100} & \textbf{100} & \textbf{95} \\
\bottomrule
\end{tabular}
\vspace{-0.1cm}
\end{table}

\begin{figure*}[!ht]  
    \centering
    \includegraphics[width=\linewidth]{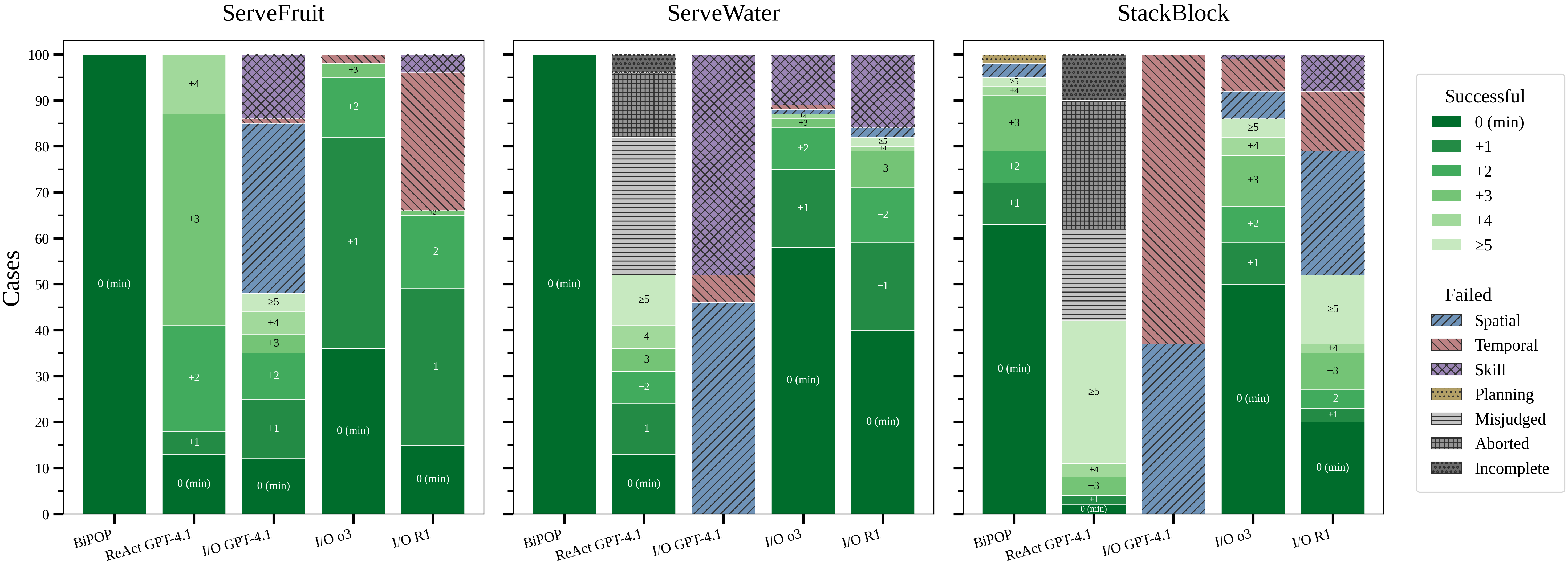}
    \caption{Per-method outcome breakdown on ServeFruit, ServeWater, and StackBlock. Each bar aggregates 100 instances: successes (green) are graded by \emph{\textbf{Optimality Gap}} ($0$, $+1,\dots,\geq 5$; dark = optimal), and failures are coloured by failure class as shown in the legend.}
    
    \label{fig:bar_all_tasks}
\end{figure*}

\noindent\textbf{Success Rate and Failure Modes.} Table~\ref{tab:success_rate} gives the per-domain success rate, and Fig.~\ref{fig:bar_all_tasks} splits all instances into successes (graded by optimality gap) and failures by class. (1) BiPOP matches or exceeds every baseline in all domains and is the only method to reach at least 95\% success throughout; although built on the non-reasoning GPT-4.1, it far surpasses I/O planning with the same model, thanks to coupling LLM coding and temporal planning with pre-execution validation. (2) These long-horizon bimanual tasks demand strong reasoning: the reasoning I/O models (o3, R1) recover much of GPT-4.1's gap, yet no LLM-only baseline matches the planner-coupled approach. (3) We observe diverse failure modes. \textit{I/O agents} exhibit \textit{Spatial} errors (out-of-reach targets, untracked staging/serve-point occupancy), \textit{Temporal} errors (both hands to one point, grasping a buried block, releasing before placement, wrong stacking order), and \textit{Skill} errors (mis-called primitives, e.g., a single-hand grasp of the co-hold bowl, or wrong arguments). \textit{BiPOP} shows the same \textit{Spatial} errors but far more rarely, and these arise from the LLM-authored PDDL specification rather than execution (mis-specified serve point), plus \textit{Planning} failures where the TFD planner times out, all confined to StackBlock and easily auditable. \textit{ReAct} failures are \textit{Misjudged} (declares completion while the goal is unmet, repeating the I/O mistakes undetected), \textit{Aborted} (detects a deadlock and gives up), and \textit{Incomplete} (exhausts its step budget\footnote{Following the I/O agents' plan lengths, the ReAct step budgets are 20, 20, and 40 for ServeFruit, ServeWater, and StackBlock, respectively.}). Overall, LLM-only planning methods fail to ensure physical-action feasibility, whereas coupling the LLM with a planner shifts the small residual error to the transparent specification and planning stages.

\noindent\textbf{Optimality Gap.} The green portion of Fig.~\ref{fig:bar_all_tasks} grades each success by its OG value (darkest = optimal). (1) BiPOP concentrates at the minimum on ServeFruit and ServeWater, almost always producing the most efficient plan. (2) On StackBlock the larger search space prevents the TFD planner from always converging within the time budget, spreading its OG beyond zero, yet it stays far more reliable than the baselines. (3) The I/O agents occasionally beat BiPOP on StackBlock, but only at a much lower overall success rate. (4) ReAct, even when successful, uses markedly longer symbolic makespans, confirming that closed-loop retrying buys coverage at the cost of parsimony.

\subsection{Discussion}
Our design separates the contribution of the planner from raw LLM capability: the I/O baselines receive the same spatio-temporal knowledge as BiPOP, in natural language rather than as a PDDL domain. The gap to BiPOP thus reflects delegating constraint enforcement and concurrency scheduling to the planner rather than model differences, and the residual gap of o3 and R1 shows that scaling reasoning alone does not close it. We accordingly build BiPOP on the non-reasoning GPT-4.1: problem generation is a constrained translation task at which it already attains near-ceiling success, whereas a reasoning backbone would add an order of magnitude of latency with little room for improvement.

To ensure comparability, we prompt I/O and ReAct for a valid plan with the fewest parallel steps. Under this objective, the reasoning models attempt aggressive parallelization, and most of their failures are spatial or temporal violations of the coordination constraints. BiPOP avoids this failure mode by construction: with the constraints in the PDDL domain, the LLM only grounds objects to reachability regions and emits a well-formed problem, while the planner handles arm assignment and makespan-minimizing concurrency, shifting the principal challenge to robust natural-language-to-PDDL alignment.

The reasoning baselines show that LLMs can exploit this spatio-temporal knowledge, reaching near-optimal plans in favourable configurations, yet hallucination and opacity still yield errors that go undetected without external verification. Symbolic planners can aid reasoning models: once a bimanual task is expressed in PDDL, plans proposed by reasoning models can be pre-verified with VAL, screening efficient but error-prone candidates before execution.

\subsection{Real-World Validation}
Since BiPOP operates at the task-planning level, the sim-to-real gap relevant to our contribution lies in perception: the pipeline must recover correct spatial relations and produce valid plans accordingly. The perception front-end is lightweight and detector-agnostic: LocateAnything \cite{wang2025locateanything} localizes the task-relevant objects, and a single camera-to-table calibration maps bounding boxes to metric coordinates, assigning each object unambiguously to the robot's workspace region. In the real-world setup, ServeWater uses a glass and a bottle instead of the two cups, and StackBlock uses a grey tray instead of the brown box. Evaluated on 100 real images per task domain with varied object configurations, the front-end localizes all task-relevant objects correctly on ServeFruit and ServeWater. On StackBlock, errors are limited to occasional mislocalizations of the grey tray and, more often, of vertically stacked blocks in the initial configuration. They stem from the detector and the fixed eye-camera viewpoint rather than from the planning framework, and they are method-agnostic: BiPOP and all baselines consume the same scene description, so perception quality shifts the shared input rather than the comparison. The decoupling also works in our favour: a stronger open-vocabulary detector or an added viewpoint upgrades the whole pipeline without touching the planner. With this front-end, BiPOP completes the ServeWater task end to end on the real robot: across varied task configurations, it produces correct partial-order plans, and executes them in live human-robot interaction through hand-crafted atomic skills. Transfer thus requires minimal additional engineering, with low-level execution left to the motion layer.

\section{Conclusion}

We present BiPOP, a task-planning framework for bimanual manipulation that combines the coding and reasoning abilities of LLMs with temporal partial-order planning. An LLM translates scene information and natural-language task descriptions into PDDL, and a temporal planner computes efficient concurrent plans whose correctness is guaranteed at the symbolic level. Across three task domains, BiPOP attains higher success rates and more efficient plans at shorter planning time than strong reasoning models, and the full pipeline transfers to the real bimanual robot NICOL.

\noindent\textbf{Limitations and Future Work.} BiPOP builds on low-level motion primitives, and its decomposition of tasks into reusable unimanual skills and a few synchronous bimanual ones facilitates learning them. Because action effects are explicit in PDDL, execution monitoring and replanning can be added to recover from failures such as unsuccessful grasps. Motion-layer duration estimates can replace the uniform durations to obtain tighter schedules. Finally, a single domain file already covers all three tasks in simulation and on the real robot, and inducing it from skill libraries or demonstrations would support open-ended skill repertoires.



\bibliographystyle{IEEEtran}
\bibliography{IEEEabrv, references}

\end{document}